\documentclass[letterpaper]{article}
\usepackage{aaai}
\usepackage{times}
\usepackage{helvet}
\usepackage{courier}
\usepackage{graphicx}
\usepackage{amsmath}
\usepackage[ruled,vlined]{algorithm2e}
\usepackage{amssymb}
\usepackage{multirow}

\newcommand{\A}{{\mathcal A}}

\newtheorem{definition}{Definition}

\begin{document}
%
\title{Game-Theoretic Approach for Non-Cooperative Planning}
\author{ 
Jaume Jord\'{a}n \and Eva Onaindia\\
Universitat Polit\`{e}cnica de Val\`{e}ncia\\
Departamento de Sistemas Inform\'{a}ticos y Computaci\'{o}n\\
Camino de Vera s/n. 46022 Valencia, Spain\\
\{jjordan,onaindia\}@dsic.upv.es
}
\maketitle
\begin{abstract}
\begin{quote}

When two or more self-interested agents put their plans to execution in the same environment, conflicts may arise as a consequence, for instance, of a common utilization of resources. In this case, an agent can postpone the execution of a particular action, if this punctually solves the conflict, or it can resort to execute a different plan if the agent's payoff significantly diminishes due to the action deferral. In this paper, we present a game-theoretic approach to non-cooperative planning that helps predict before execution what plan schedules agents will adopt so that the set of strategies of all agents constitute a Nash equilibrium. We perform some experiments and discuss the solutions obtained with our game-theoretical approach, analyzing how the conflicts between the plans determine the strategic behavior of the agents.

\end{quote}
\end{abstract}

\section{Introduction}

Multi-agent Planning (MAP) with self-interested agents is the problem of coordinating a group of agents that compete to make their strategic behavior prevail over the others': agents competing for a particular goal or the utilization of a common resource, agents competing to maximize their benefit or agents willing to form coalitions with others in order to achieve better their own goals or preferences. In this paper, we focus on game-theoretic MAP approaches for self-interested agents.

Brafman et al \cite{BrafmanDET09} introduce the Coalition-Planning Game (CoPG), a game-theoretic approach for self-interested agents which have personal goals and costs but may find it beneficial to cooperate with each other provided that the coalition formation helps increase their personal net benefit. In particular, authors propose a theoretical framework for stable planning in acyclic CoPG which is limited to one goal per agent. Following the line of CoPG, the work in \cite{CrosbyR11} presents an approach that combines heuristic calculations in existing planners for solving a restricted subset of CoPGs. In general, there has been a rather intensive research on cooperative self-interest agents as, for example, for modeling the behavior of planning agents in groups \cite{Hadad13} and in coalitional resource game scenarios \cite{Dunne2010}, among others.

On the other hand, game-theoretic non-cooperative MAP approaches aim, in general, at finding a Nash Equilibrium joint plan out of the individual plans of the agents. 'Pure' game-theoretic approaches, like \cite{BowlingJV03} and \cite{LarbiKM07} perform a strategic analysis of all possible agent plans and define notions of equilibria by analyzing the relationships between different solutions in game-theoretic terms. In \cite{BowlingJV03}, MAP solutions are classified according to the agents' possibility of reaching their goals and the paths of execution (combinations of local plans). Similarly, satisfaction profiles in \cite{LarbiKM07} are defined by the level of assurance of reaching the agent's goals. A different approach using best-response was proposed to solve congestion games and to perform plan improvement in general MAP scenarios from an available initial joint plan \cite{JonssonR11}.

Game-theoretic approaches that evaluate every strategy of every agent against all other strategies are ineffective for planning,
since even if plan length is bounded polynomially, the number of available strategies is exponential \cite{Nissim13}. However, in environments where cooperation is not allowed or calculating an initial joint plan is not possible, game-theoretic approaches are useful. Take, for instance, the modeling of a transportation network, sending packets through the Internet or a network traffic, where individuals need to evaluate routes in the presence of the congestion resulting from the decisions made by themselves and everyone else \cite{EasleyK10}. In this sense, we argue that game-theoretic reasoning is a valid approach for this specific type of planning problems, among others.

In this paper, we present a novel game-theoretic non-cooperative model to MAP with self-interested agents that solves the following problem. We consider a group of agents where each agent has one or several plans that achieve one or more goals. Executing a particular plan reports a benefit to the agent depending on the number of goals achieved, makespan of the plan or cost of the actions. Agents operate in a common environment, what may provoke interactions between the agents' plans and thus preventing a concurrent execution. Each agent is willing to execute the plan that maximizes its benefit but it ignores which plan the other agents will point out, how his plan will be interleaved with theirs and the impact of such coordination on his benefit.

We present a two-game proposal to tackle this problem. A \emph{general game} in which agents take a strategic decision on which joint plan to execute, and an \emph{internal game} that, given one plan per agent, returns an equilibrium joint plan schedule. Agents play the internal game to simulate the simultaneous execution of their plans, find out the possibilities to coordinate in case of interactions and the effect of such coordination on their final benefit. The approach of the general game is very similar to
the work described in \cite{LarbiKM07}; specifically, our proposal contributes with several novelties:

\begin{itemize}
\item Introduction of \emph{soft goals} to account for the case in which a joint plan that achieves all the goals of every agent is not feasible due to the interactions between the agents' plans. The aim of the general game is precisely to select an equilibrium joint plan that encompasses the 'best' plan of each agent.
\item  An explicit handling of conflicts between actions and a mechanism for updating the plan benefit based on the penalty derived from the conflict repair. This is precisely the objective of the internal game and the key contribution that makes our model a more realistic approach to MAP with self-interested agents.
\item An implementation of the theoretical framework, using the Gambit tool \cite{Gambit} for solving the general game and our own program for the internal game.
\end{itemize}

We wish to highlight that the model presented in this paper is not intended to solve a complete planning problem due to the exponential complexity inherent to game-theoretic approaches. The model is aimed at solving a specific situation where the alternative plans of the agents are particularly limited to such situation and thus plans would be of a relatively similar and small size.

The paper is organized as follows. The next section provides an overview of the problem, introduces the notation that we will use throughout the paper and describes the general game in detail. The following section is devoted to the specification of the internal game, which we call the joint plan schedule game. Section 'Experimental results' shows some experiments carried out with our model and last section concludes.

\section{Problem Specification}

The problem we want to solve is specified as follows. There is a set of $n$ rational, self-interested agents $N=\{1,..., n\}$ where each agent $i$ has a collection of independent plans $\Pi_i$ that accomplish one or several goals. Executing a particular plan $\pi$ provides the owner agent a real-valued benefit given by the function $\beta: \Pi \rightarrow \mathbb{R}$. The benefit that agent $i$ obtains from plan $\pi$ is denoted by $\beta_i(\pi)$; in this work, we make this value dependent on the number of goals achieved by $\pi$ and the makespan of $\pi$ but different measures of reward and cost might be used, like the relevance of the achieved goals to agent $i$ or the cost of the actions of $\pi$. Each agent $i$ wishes to execute a plan $\pi$ such that $max(\beta_i(\pi)), \forall \pi \in \Pi_i$; however, since agents have to execute their plans simultaneously in a common environment, conflicts may arise that prevent agents from executing their preferable plans. Let's assume that $\pi$ and $\pi^\prime$ are the maximum benefit plans of agents $i$ and $j$, respectively, and that the simultaneous execution of $\pi$ and $\pi^\prime$ is not possible due to a conflict between the two plans. If this happens, several options are analyzed:

\begin{itemize}
\item agent $i$ (agent $j$, respectively) considers to \emph{adapt} the execution of its plan $\pi$ ($\pi^\prime$, respectively) to the plan of the other agent by, for instance, delaying the execution of one or more actions of $\pi$ so that this delay solves the conflict. This has an impact in $\beta_i(\pi)$ since any delay in the execution of $\pi$ diminishes the value of its original benefit.
\item agent $i$ (agent $j$, respectively) considers to switch to another plan in $\Pi_i$ ($\Pi_j$, respectively) which does not cause any conflict with the plan $\pi^\prime$ ($\pi$, respectively).
\end{itemize}

Agents wish to choose their maximum benefit plan but then the choices of the other agents can affect each other's benefits. This is the reason we propose a game-theoretic approach to solve this problem.

A plan $\pi$ is defined as a sequence of non-temporal actions $\pi =[ a_1, \ldots, a_m ]$\footnote{In this first approach, we consider only instantaneous actions}. Assuming $t=0$ is the start time when the agents begin the execution of one of their plans, the execution of $\pi$ would ideally take $m$ units of time, executing $a_1$ at time $t=0$ and the rest of actions at consecutive time instants, thus finishing the execution of $\pi$ at time $t=m-1$ (last action is scheduled at $m-1$). This is called the \emph{earliest plan execution} as it denotes that the start time and finish time of the execution $\pi$ are scheduled at the earliest possible times. However, if conflicts between $\pi$ and the plans of other agents arise, then the actions of $\pi$ might likely not to be realized at their earliest times, in which case a tentative solution could be to delay the execution of some action in $\pi$ so as to avoid the conflict. Therefore, given a plan $\pi$, we can find infinite schedules for the execution of $\pi$.

\begin{definition} Given a plan $\pi =[ a_1, \ldots, a_m ]$,  $\Psi_{\pi}$ is an infinite set that contains all possible schedules for $\pi$. Particularly, we define as $\psi_0$ the earliest plan execution of $\pi$ that finishes at time $m-1$. Given two different schedules $\psi_j, \psi_{j+1} \in \Psi_{\pi}$, the finish time of $\psi_j$ is prior or equal to the finish time of $\psi_{j+1}$.
\end{definition}

Let $\psi_j$ , where $j \not= 0$, be a schedule for $\pi$ that finishes at time $t > m-1$. The net benefit that the agent obtains with $\psi_j$ diminishes with respect to $\beta_i(\pi)$. The loss of benefit is a consequence of the delayed execution of $\pi$ and this delay may affect agents differently. For instance, if for agent $i$ the delay of $\psi_j$ wrt to $\psi_0$ has a low incidence in $\beta_i(\pi)$, then $i$ might still wish to execute $\psi_j$. However, for a different agent $k$, a particular schedule of
a plan $\pi^\prime \in \Pi_k$  may have a great impact in $\beta_k(\pi^\prime)$ even resulting in a negative net benefit. How delays affect the benefit of the agents depends on the intrinsic characteristics of the agents.

\begin{definition} We define a utility function $\mu: \Psi \rightarrow \mathbb{R}$ that returns the net value of a plan schedule. Thus, $\mu_i (\psi_j)$, $\psi_j \in \Psi_{\pi}$, is the utility that agent $i$ receives from executing the schedule $\psi_j$ for plan $\pi$. By default, for any given plan $\pi$ and $\psi_0 \in \Psi_{\pi}$, $\mu_i (\psi_0) = \beta_i(\pi)$.
\end{definition}

A rational way of solving the conflicts of interest that arise among a set of self-interested agents who all wish to execute their maximum benefit plan comes from the non-cooperative game theory. Therefore, our general game is modeled as a non-cooperative game in the \emph{Normal-Form}. The agents are the players of the game; the set of actions $\A_i$ is modeled as the game actions (plans) available to agent $i$, and the payoff function is defined as the result of a rational selection of a plan schedule for each agent. Formally:

\begin{definition} We define our general game as a tuple ($N, P, \rho$), where:
\begin{itemize}
	\item $N=\{1, \ldots, n\}$ is the set of $n$ self-interested players.
	\item $P=P_1 \times ... \times P_n$, where $P_i=\Pi_i, \forall i \in N$. Each agent $i$ has a finite set of strategies which are the plans contained in $\Pi_i$. We will then call a \emph{plan profile} the n-tuple $p=(p_1, p_2, \ldots, p_n)$, where $ p_i \in \Pi_i$ for each agent $i$.
	\item $\rho=(\rho_1,...,\rho_n)$ where $\rho_i: P \rightarrow \mathbb{R}$ is a real-valued payoff function for agent $i$. $\rho_i(p)$ is defined as the utility of the schedule of plan $p_i$ when $p_i$ is executed simultaneously with $(p_1, \ldots, p_{i-1}, p_{i+1}, \ldots, p_n)$.
\end{itemize}
\end{definition}

The plan profile $p$ represents the plan choice of each agent. Every agent $i$ wishes to execute the schedule $\psi_0 \in \Psi_{p_i}$. Since this may not be feasible, agents have to agree on a joint plan schedule. We define a procedure named \emph{joint plan schedule} that receives as input a plan profile $p$ and returns a schedule profile $s=(s_1, s_2, \ldots, s_n)$, where $\forall i \in N, s_i \in \Psi_{p_i}$. The schedule profile $s$ is a consistent joint plan schedule; i.e., all of the individual plan schedules in $s$ can be simultaneously executed without provoking any conflict. The \emph{joint plan schedule} procedure, whose details are given in the next section, defines our internal game.

Let $p=(p_1, p_2, \ldots, p_n)$ be a plan profile and $s=(s_1, s_2, \ldots, s_n)$ the schedule profile for $p$. Then, we have that $\rho_i(p)=\mu_i(s_i)$.

The game returns a scheduled plan profile that is a Nash Equilibrium (NE) solution. This represents a stable solution from which no agent benefits from invalidating another agent's plan schedule.

\section{The \emph{joint plan schedule} game}

This section describes the internal game. The problem consists in finding a feasible joint plan schedule for a given plan profile $p=(p_1, p_2, \ldots, p_n)$, where each agent $i$ wishes to execute its plan $p_i$ under the earliest plan schedule ($\psi_0$). Since potential conflicts between the actions of the plans of different agents may prevent some of them from executing $\psi_0$, agents get engaged in a game in order to come up with a rational decision that maximizes their expected utility.

For a particular plan $\pi$, an action $a \in \pi$ is given by the triple $a=\langle pre(a),add(a),del(a) \rangle$, where $pre(a)$ is the set of conditions that must hold in a state $S$ for the action to be applicable, $add(a)$ is its add list, and $del(a)$ is its delete list, each a set of literals. Let $a$ and $a^\prime$ be two actions, both scheduled at time $t$, in the plans of two different agents; a conflict between $a$ and $a^\prime$ occurs at $t$ if the two actions are mutually exclusive (mutex) at $t$ \cite{BlumF97}.

The joint plan schedule game is actually the result of simulating the execution of all the agents' plans. At each time $t$, every agent $i$ makes a move, which consists in executing the next action $a$ in its plan $p_i$ or executing the empty action ($\bot$). The empty action is the default mechanism to avoid two actions that are mutex at $t$, and this implies a deferral in the execution of $a$. A concept similar to $\bot$, called the empty sequence, is used in \cite{LarbiKM07} as a neutral element for calculating the permutations of the plans of two agents, although the particular implication of this empty sequence in the plan or in the evaluation of the satisfaction profiles is not described.

\subsection{Search space of the internal game}

Several issues must be considered when creating the search space of the internal game:

\begin{itemize}

\item[1)] \textbf{Simultaneous and sequential execution of the game}. The internal game is essentially a multi-round sequential game since the simulation of the plans execution occurs along time, one action of each player at a time. Then, the execution at time $t+1$ only takes place when every agent has moved at time $t$, so that players observe the choices of the rest of agents at $t$. In contrast, the game at time $t$ represents the simultaneous moves of the agents at that time. Simultaneous moves can always be rephrased as sequential moves with imperfect information, in which case agents would likely get 'stuck' if their actions are mutex; that is, agents would not have the possibility of coordinating their actions. Therefore, simultaneous moves at $t$ are also simulated as sequential moves as if agents would know the intention of the other agents. In essence, this can be interpreted as agents analyzing the possibilities of avoiding the conflict and then playing simultaneously the choice that reports a stable solution. Obviously, this means that agents would know the strategies of the others at time $t$, what seems reasonable if they are all interested in maximizing their utility.

\item [2)] \textbf{Applicability of the actions}. Unlike other games where the agents' strategies are always \emph{applicable}, in planning it may happen that an action $a$ of a plan is not executable at time $t$ in the state resulting from the execution of the $t-1$ previous steps. In such a case, the schedule profile is discarded. In our model, a schedule profile $s$ is a solution if $s$ comprises a plan schedule for every agent. Otherwise, we would be considering coalitions of agents that discard strategies that do not \emph{fit} with the strategies of the coalition members. On the other hand, $\bot$ is only applicable at $t$ if at least any other agent applies a non-empty action at $t$ . The empty action is also applicable when the agent has played all the actions of its plan.

\end{itemize}

\textbf{Example}. Consider a plan profile $p=(p_1, p_2)$ of two agents, where $p_1 =[ a_1, a_2, a_3]$ and $p_2 =[ b_1, b_2, b_3]$. $s=(s_1, s_2)$ with $s_1=(a_1, \bot, \bot, a_2, a_3)$ and $s_2=(\bot, b_1, b_2, b_3, \bot)$ is a valid joint schedule if all the actions scheduled at each time $t$ are not mutex.

\begin{definition} Given a plan profile $p=(p_1, p_2, \ldots, p_n)$, $s=(s_1, s_2, \ldots, s_n)$ is a valid schedule profile to $p$
if every $s_i$ is a non-empty plan schedule and the actions of every $p_i$ scheduled at each time $t$ are not mutex.
\end{definition}

Following, we formally define our internal game.

\begin{definition} A perfect-information extensive-form game consists of:
\begin{itemize}
	\item a set of players, $N=\{1, \ldots, n\}$
	\item a finite set $X$ of nodes that form the tree, with $S \subset X$ being the terminal nodes
	\item a set of functions that describe each $x \not\in S$:
        \begin{itemize}
            \item the player $i(x)$ who moves at $x$
            \item the set $A(x)$ of possible actions at $x$
            \item the successor node $n(x,a)$ resulting from action $a$
        \end{itemize}
    \item $n$ payoff functions that assign a payoff to each player as a function of the terminal node reached
\end{itemize}
\end{definition}

Let $p_i=[a_1,\ldots,a_m]$ be the plan of agent $i$. The set $A(x)$ of possible actions of $i$ at $x$ is $A(x)=\{a, \bot\}$, where
$a$ is the action of $p_i$ that has to be executed next, which comes determined by the evolution of the game so far. Only in the case that agent $i$ has already played the $m$ actions of $p_i$, $A(x)=\{\bot\}$. As commented above, each agent makes a move at a time so the first $n$ levels of the tree represent the moves of the $n$ agents at time $t$, the next $n$ levels represent the moves of the $n$ agents at $t+1$ and so on.

A node $x$ of the game tree represents the planning state after executing the path from the root node until $x$. For each node $x$,  there are at most two successor nodes, each corresponding to the application of the actions in $A(x)$. A terminal node $s$ denotes a valid schedule profile.

Let $s=(s_1, s_2, \ldots, s_n)$ be a terminal node; the payoff of player $i$ at $s$ is given by $\mu_i(s_i)$. Note that the solution of the internal game for a plan profile $p=(p_1, p_2, \ldots, p_n)$ is one of the terminal nodes of the game tree, and the payoff for each player $i$ represents the value of $\rho_i(p)$. Then, the payoff vector of the solution terminal node is the payoff vector of one of the cells in the general game.

\subsection{Subgame Perfect Equilibrium (SPE)}

The solution concept we apply in our internal game is the Subgame Perfect Equilibrium \cite[Chapter 5]{Shoham09}, a concept that refines a NE in perfect-information extensive-form games by eliminating those unwanted Nash Equilibra. The SPE of a game are all strategy profiles that are NE for any subgame. By definition, every SPE is also a NE, but not every NE is SPE. The SPE eliminates the so-called ``noncredible threats'', that is, those situations in which an agent $i$ threatens the other agents to choose a node that is harmful for all of them, with the intention of forcing the other players to change their decisions, thus allowing $i$ to reach a more profitable node. However, this type of threats are non credible because a self-interested agent would not jeopardize its utility.

A common method to find a SPE in a finite perfect-information extensive-form game is the backward induction algorithm. This algorithm has the advantage that it can be computed in linear time in the size of the game tree, in contrast to the best known methods to find NE that require time exponential in the size of the normal-form. In addition, it can be implemented as a single depth-first traversal of the game tree. We consider the SPE as the most adequate solution concept for our joint plan schedule game since SPE reflects the strategic behavior of a self-interested agent taking into account the decision of the rest of agents to reach the most preferable solution in a common environment.

The SPE solution concept provides us a strong argument to solve the problem of selecting a joint plan schedule as a perfect-information extensive-form game instead of using, for example, a planner that returns all possible combinations of the agents' plans. In this latter case, the question would be which policy to apply to choose one schedule over the other. We could apply criteria such as Pareto-optimality\footnote{A vector Pareto-dominates another one if each of the components of the first one is greater or equal to the corresponding component in the second one.} or the maximum social welfare\footnote{The sum of all agents' utility}. However, a Pareto-dominant solution does not always exist in all problems and the highest social welfare solution may be different from the SPE solution. That is, neither of these solution concepts would actually reflect how the fate of one agent is impacted by the actions of others.

The SPE solution concept has also some limitations. First, there could exist multiple SPE in a game, in which case one SPE may be chosen randomly. Second, the order of the agents when building the tree is relevant for the game in some situations. Consider, for instance, the case of a two-agent game. The application of the backward induction algorithm would give some advantage to the first agent in those cases for which there exist two different schedules to avoid the mutex (delaying one agent's action over the other or viceversa). In this case, the first agent will then select the solution that does not delay its conflicting action. Notice that in these situations both solutions are SPE and thus equally good from a game-theoretic perspective. Any other conflict-solving mechanism would also favour one agent over the other one depending on the used criteria; for instance, a planner would favour the agent whose delay returns the shortest makespan solution, and a more social-oriented approach would give advantage to the agent whose delay minimizes the overall welfare. In order to alleviate the impact of the order of the agents in the SPE solution, agents are randomly chosen in the tree generation.

\begin{figure}
    \centering
    \includegraphics[width=0.45\textwidth]{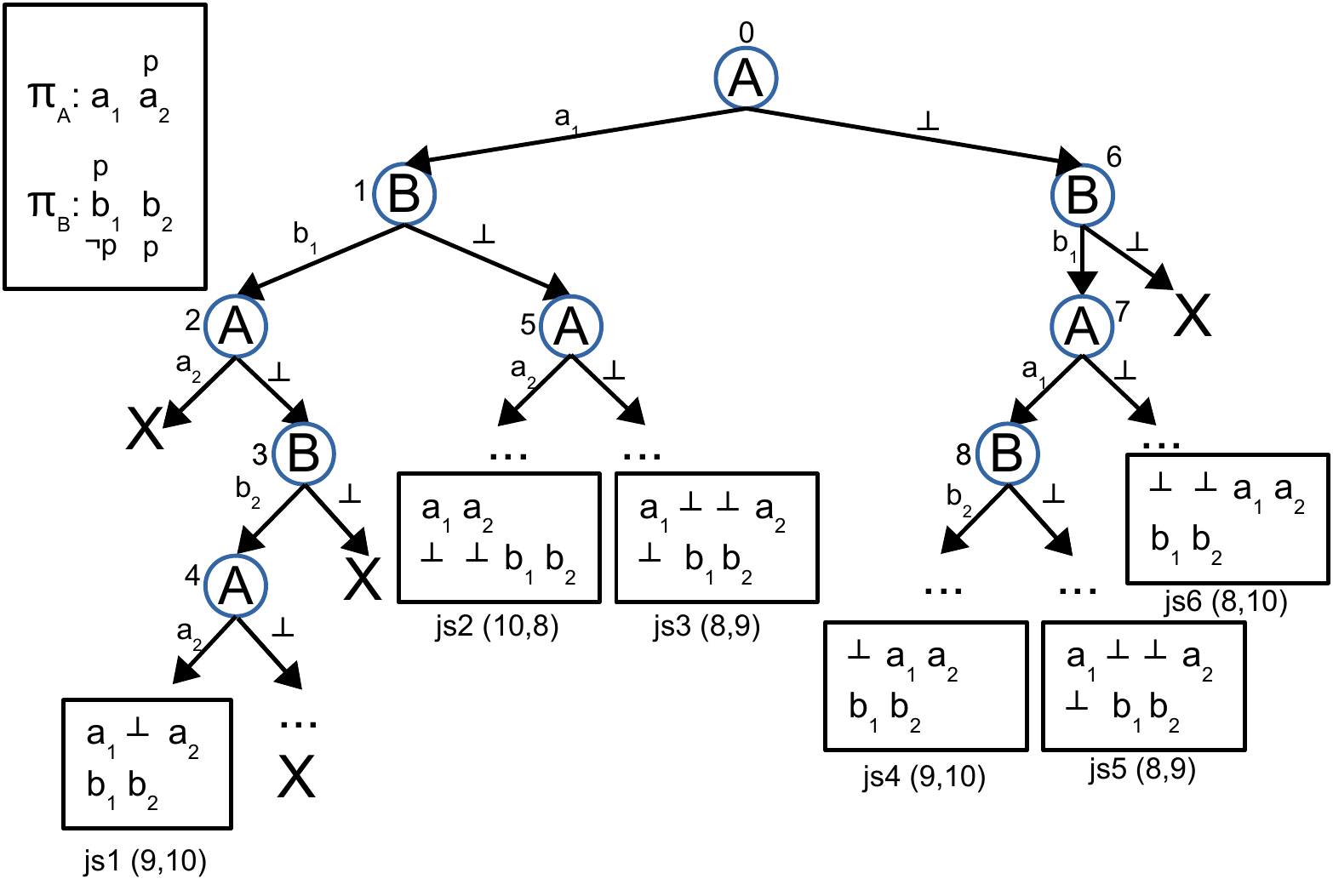}
    \caption{Tree example}
    \label{fig:tree}
\end{figure}

An example of an extensive-form tree for a particular joint plan schedule problem can be seen in Figure \ref{fig:tree}. The tree represents the internal game of two agents A and B with plans $\pi_A=[a_1, a_2]$ and $\pi_B=[b_1, b_2]$. The letter above an action represent its precondition, the letter below represents its effects. Thus, $p \in pre(a_2)$, $p \in pre(b_1)$, $\neg p \in del(b_1)$ and $p \in add(b_2)$. At each non-terminal node, the corresponding agent generates its successors; in case of a non-applicable action, the branch is pruned. For example, in node 2 agent A tries to put its action $a_2$, but this is not possible because in that state a previous action $b_1$ deleted $p$. Another example of non applicable action is shown in the right branch of node 6. In this case, agent B tries to apply the empty action $\bot$, but this option is also discarded because agent A has also applied an empty action in the same time step ($t=0$).

In the tree example of Figure \ref{fig:tree}, we assume that both agents A and B have the same utility function, that a delay means a penalty proportional to the utility, and that $\beta(\pi_A) = \beta(\pi_B) = 10$. If we apply the backward induction algorithm to the this extensive-form game, it returns the joint schedule profile $js1$, or its equivalent $js4$. This schedule profile reports the highest possible utility for agent B, and a penalty of one unit (generic penalty) for agent A. Let's see how the backward induction algorithm obtains the SPE in this example. The payoffs of $js1$ are back up to node 2, where they will be compared with the values of node 5.  The joint schedule $js2$ is backed up to node 5 because agent A is who chooses at node 5. Then, in node 1 agent B chooses between node 2 and node 5 and hence, $js1$ is chosen. In the other branches, in node 8 $js4$ will prevail over $js5$ and then, when compared in node 7 with $js6$, the choice of agent A is $js4$. This results in agent A choosing at node 0 between $js1$ and $js4$, both with the same payoffs, and so both are equivalent SPE solutions. If the tree is developed following a different agent order the SPE solution will be the same.

\section{Experimental results}
\label{experiments}

In this section, we present some experimental results in order to validate and discuss our approach. As several factors can affect the solutions of the general game, we show different examples of game situations. 

We implemented a program to generation the extensive-form tree and apply the backward induction algorithm. The NE in the normal-form game is computed with the tool Gambit \cite{Gambit}.

For the experiments we used problems of the well-known Zeno-Travel domain from the International Planning Competition (IPC-3)\footnote{http://ipc.icaps-conference.org/}. However, for simplicity and the sake of clarity, we show generic actions in the figures. 

The experiments were carried out for two agents, A and B. Both agents have a set of individual plans that solve one or more goals. The more goals achieved by a plan, the more the benefit of the plan. In addition, the benefit of a plan depends on the makespan of such plan. Given a plan $\pi$, which earliest plan execution is denoted by $\psi_0$, $\beta_i(\pi)$ is calculated as follows: $\beta_i(\pi)=nGoals(\pi)*10 - makespan(\psi_0)$, where $nGoals(\pi)$ represents the number of goals solved by $\pi$ and $makespan(\psi_0)$ represents the minimum duration schedule for $\pi$.

The utility of a particular schedule $\psi \in \Psi_{\pi}$ is a function of $\beta_i(\pi)$ and the number of time units that the actions of $\pi$ are delayed in $\psi$ with respect to the earliest plan execution $\psi_0$; in other words, the difference in the makespan of $\psi$ and $\psi_0$. Thus, $\mu_i(\psi)=\beta_i(\pi)$ if $\psi = \psi_0$. Otherwise, $\mu_i(\psi)=\beta_i(\pi)- delay(\psi)$, where $delay(\psi)$ is the delay in the makespan of $\psi$ with respect to the makespan of $\psi_0$.

\begin{table}[ht]
\centering \footnotesize
\begin{tabular}{|c|c|l|l|l|}
 \hline
 Problem & Agent  & Plan & nAct($\pi$) & $\beta_i(\pi)$ \\ \hline

\multirow{6}{*}{1} & \multirow{4}{*}{A} &  $\pi_{A1}(g_1g_2)$ & 3 &  17  \\
 & &   $\pi_{A2}(g_1)$ & 2 & 8 \\
 & &   $\pi_{A3}(g_2)$ & 1 & 9 \\  \cline{2-5}
 & \multirow{3}{*}{B} &  $\pi_{B1}(g_1g_2)$ & 2 &  18  \\
 & &  $\pi_{B2}(g_1)$ & 1 & 9 \\
 & & $\pi_{B3}(g_2)$ & 1 & 9 \\

 \hline \hline

 \multirow{8}{*}{2} & \multirow{4}{*}{A} &  $\pi_{A1}(g_1g_2)$ & 2 &  18  \\
 & &  $\pi_{A2}(g_1g_2)$ & 4 & 16 \\
 & &   $\pi_{A3}(g_1)$ & 2 & 8 \\
 & &   $\pi_{A4}(g_2)$ & 1 & 9 \\  \cline{2-5}
 & \multirow{3}{*}{B} &  $\pi_{B1}(g_1g_2)$ & 2 &  18  \\
 & &  $\pi_{B2}(g_1g_2)$ & 4 & 16 \\
 & &  $\pi_{B3}(g_1)$ & 1 & 9 \\
 & & $\pi_{B4}(g_2)$ & 1 & 9 \\ \hline

 \end{tabular}
\caption{Problems description}
\label{tab:setup}
\end{table}

Table \ref{tab:setup} shows the problems used in these experiments: the set of initial plans of each agent, the number of actions of each plan and its utility.

\begin{table}[ht]
\centering \footnotesize
\begin{tabular}{|l|c|c|c|}
 \hline

	&  $\pi_{B1}$ & $\pi_{B2}$ &  $\pi_{B3}$ \\ \hline
$\pi_{A1}$ &  15,16 (2,2) &  17,7 (0,2)  & 17,9 (0,0) \\ \hline
$\pi_{A2}$  & 8,16 (0,2) & 8,7 (0,2) & 7,9 (1,0)\\ \hline
$\pi_{A3}$ &  7,18 (2,0)&  9,9 (0,0) & 8,9 (1,0) \\ \hline

\end{tabular}
\caption{Problem 1}
\label{tab:prob1}
\end{table}

In Table \ref{tab:prob1} we can see the results of the general game for problem 1. Each cell is the result of a joint plan schedule game that combines a plan of agent A and a plan of agent B. In each cell, we show the payoff of $\pi_{Ax}$ and $\pi_{By}$ as well as the values of $delay(\psi)$ for each plan (delay values are shown between parenthesis). The values in each cell are the result of the schedule profile returned by the internal game.

\begin{figure}[ht]
    \centering
    \includegraphics[width=0.32\textwidth]{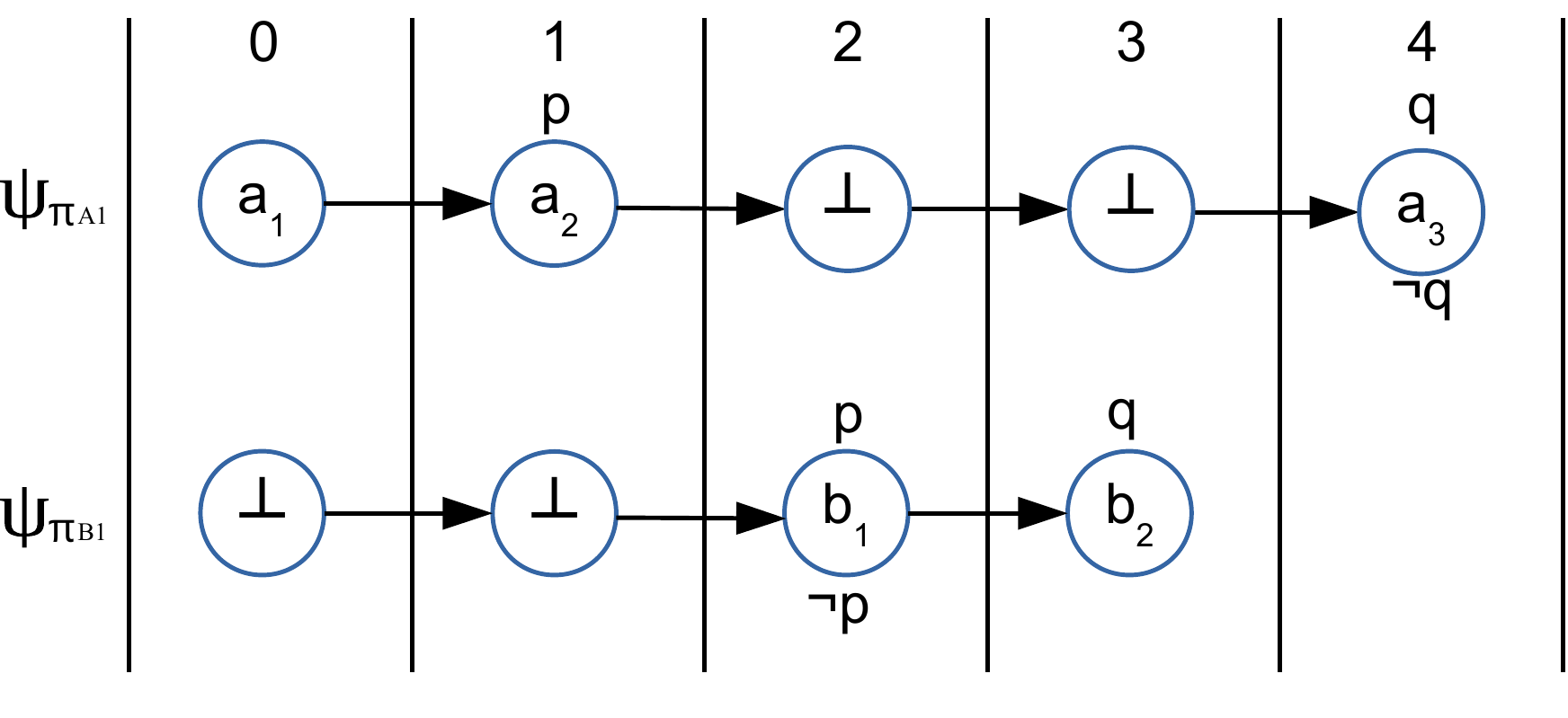}
    \caption{Schedule example}
    \label{fig:schedule}
\end{figure}

The NE of this problem is the combination of $\pi_{A1}$ and $\pi_{B1}$, with an utility of (15,16) for agent A and B, respectively. Agent A uses the plan that solves its goals $g_1$ and $g_2$ delayed two time steps. Agent B uses the plan that solves its goals $g_1$ and $g_2$, also delayed two time steps. The solution for both agents is to use the plan that solves more goals (with a higher initial benefit) a bit delayed. This can be a typical situation if there are not many conflicts and if the delay is not very punishing to the agents. The schedule of this solution is shown in Figure \ref{fig:schedule}. We can see in the figure that agent A starts the execution of its plan $\pi_{A1}$ at $t=0$, but after having scheduled its first two actions, the strategy of agent A introduces a delay of two time steps (empty actions) until it can finally execute its final action without causing a mutex with the actions of agent B. Regarding agent B, its first action in $\pi_{B1}$ is delayed two time units to avoid the conflict with agent A. In this example, both agents have a conflict with each other (both have an action which deletes a condition that the other agent needs).

\begin{figure}
    \centering
    \includegraphics[width=0.125\textwidth]{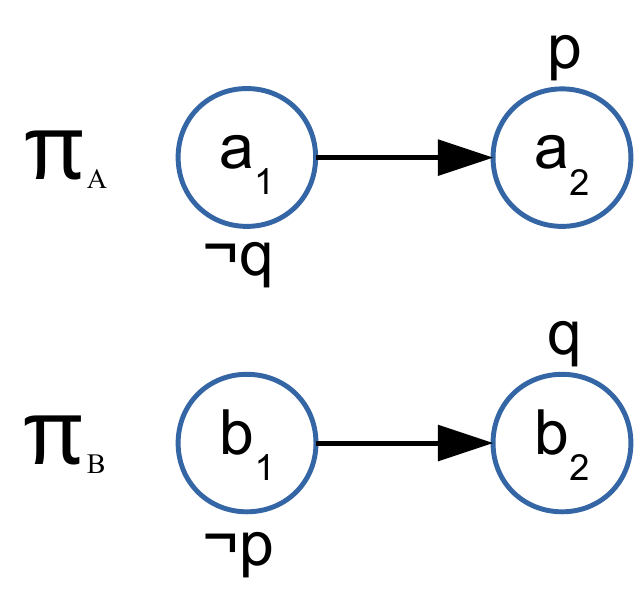}
    \caption{Unsolvable conflict}
    \label{fig:conflict}
\end{figure}

Table \ref{tab:prob2} represents the game in normal-form of problem 2 shown in Table \ref{tab:setup}. In this case, we find three different equilibria: ($\pi_{A1}$, $\pi_{B2}$) with payoffs (15,14) and delays (3,2) for agent A and B, respectively; another NE is  ($\pi_{A2}$, $\pi_{B1}$), with payoffs (14,15) and a delay of (2,3) time steps, respectively; the last NE is a mixed strategy with probabilities $0.001$ and $0.999$ for $\pi_{A1}$ and $\pi_{A2}$ of agent A, and probabilities $0.001$ and $0.999$ for strategies $\pi_{B1}$ and $\pi_{B2}$ of agent B. In this problem we have a cell with $-\infty$ as payoff of the two agents. This payoff represents that there does not exist a valid joint schedule for the plans due to an unsolvable conflict as the one shown in Figure \ref{fig:conflict}.

\begin{table}[ht]
\centering \scriptsize
\begin{tabular}{|l|c|c|c|c|}
 \hline

	&  $\pi_{B1}$ & $\pi_{B2}$ &  $\pi_{B3}$ & $\pi_{B4}$\\ \hline
$\pi_{A1}$ &  $-\infty$,$-\infty$ &  15,14 (3,2)  & 18,7 (0,2) & 17,9 (1,0) \\ \hline
$\pi_{A2}$  & 14,15 (2,3) & 14,14 (2,2) & 16,6 (0,3) & 16,9 (0,0) \\ \hline
$\pi_{A3}$ &  8,16 (0,2)&  8,16 (0,0) & 8,7 (0,2) & 8,8 (0,1) \\ \hline
$\pi_{A4}$ &  7,18 (2,0)&  6,16 (3,0) & 9,9 (0,0) & 8,9 (1,0) \\ \hline

\end{tabular}
\caption{Problem 2}
\label{tab:prob2}
\end{table}

The game in Table \ref{tab:prob2b} is the same game as the one in Table \ref{tab:prob2} but, in this case, the agents suffer a delay penalty of $3.5$ (instead of $1$) per each action delayed in their plan schedules. Under this new evaluation, we can see how this affects the general game. In this situation, the only NE solution is ($\pi_{A2}$, $\pi_{B2}$) with utility values (9,9) and a delay of two time steps for each agent. Note that this solution is neither Pareto-optimal (solution (16,9) is Pareto-optimal) nor it maximizes the social welfare. However, these two solution concepts can be applied in case of multiple NE.

\begin{table}[ht]
\centering \scriptsize
\begin{tabular}{|l|c|c|c|c|}
 \hline

	&  $\pi_{B1}$ & $\pi_{B2}$ &  $\pi_{B3}$ & $\pi_{B4}$\\ \hline
$\pi_{A1}$ &  $-\infty$,$-\infty$ &  7.5,9 (3,2)  & 18,2 (0,2) & 14.5,9 (1,0) \\ \hline
$\pi_{A2}$  & 9,7.5 (2,3) & 9,9 (2,2) & 16,-1.5 (0,3) & 16,9 (0,0) \\ \hline
$\pi_{A3}$ &  8,11 (0,2)&  8,16 (0,0) & 8,2 (0,2) & 8,5.5 (0,1) \\ \hline
$\pi_{A4}$ &  2,18 (2,0)&  -1.5,16 (3,0) & 9,9 (0,0) & 5.5,9 (1,0) \\ \hline

\end{tabular}
\caption{Problem 2b, more delay penalty to the utility}
\label{tab:prob2b}
\end{table}

In conclusion, our approach simulates how agents behave with several strategies and it returns an equilibrium solution that is stable for all of the agents. All agents participate in the schedule profile solution and their utilities are dependent on the strategies of the other agents regarding the conflicts that appear in the problem.

\section{Conclusions and future work}

In this paper, we have presented a complete game-theoretic approximation for non-cooperative agents. The strategies of the agents are determined by the different ways of solving mutex actions at a time instant and the loss of utility of the solutions in the plan schedules. We also present some experiments carried out in a particular planning domain. The results show that the SPE solution of the extensive-form game in combination with the NE of the general game return a stable solution that responds to the strategic behavior of all of the agents.

As for future work, we intend to explore two different lines of investigation. The exponential cost of this approach represents a major limitation for being used as a general MAP method for self-interested agents. Our combination of a general+internal game can be successively applied in subproblems of the agents. Considering that this approach solves a subset of goals of an agent, the agent could get engaged in a new game to solve the rest of his goals, and likewise for the rest of agents. Then, a MAP problem can be viewed as solving a subset of goals in each repetition of the whole game. In this line, the utility functions of the agents can be modeled not only to consider the benefit of the current schedule profile but also to predict the impact of this strategy profile in the resolution of the future goals. That is, we can define payoffs as a combination of the utility gained in the current game plus an estimate of how the joint plan schedule would impact in the resolution of the remaining goals.

Another line of investigation is to extend this approach to cooperative games, allowing the formation of coalitions of agents if the coalition represents a more advantageous strategy than playing \emph{alone}.

\section*{ Acknowledgments}
This work has been partly supported by the Spanish MICINN under projects Consolider Ingenio 2010 CSD2007-00022 and TIN2011-27652-C03-01, and the Valencian project PROMETEOII/2013/019.

\newpage

\bibliographystyle{aaai}
\bibliography{references}

\end{document}